\pdfoutput=1
\documentclass{article}
\usepackage[T1]{fontenc}
\usepackage{arxiv,times}

\usepackage{amsmath,amsfonts,bm}

\def\eqref#1{equation~\ref{#1}}

\def\1{\bm{1}}

\DeclareMathAlphabet{\mathsfit}{\encodingdefault}{\sfdefault}{m}{sl}
\SetMathAlphabet{\mathsfit}{bold}{\encodingdefault}{\sfdefault}{bx}{n}

\usepackage{hyperref}
\usepackage{url}
\usepackage{booktabs}
\usepackage{array}
\usepackage{graphicx}
\usepackage[caption=false,font=small]{subfig}
\usepackage{wrapfig}
\usepackage{placeins}
\usepackage{amsmath}
\usepackage{amssymb}
\usepackage{algorithm}
\usepackage{algorithmic}
\usepackage{fontawesome5}

\graphicspath{{figures/}}

\newcommand{\gsynth}{g(\tilde{x})}
\newcommand{\gref}{g_{\mathrm{ref}}}
\newcommand{\cossim}{\cos(\gsynth, \gref)}
\newcommand{\cosm}{\textsc{cos}}
\newcommand{\cosnov}{\textsc{cos}\,$\times$\,\textsc{novelty}}
\newcommand{\difficulty}{\textsc{difficulty}}
\newcommand{\solveonly}{\textsc{solve-only}}
\newcommand{\cplay}{CompassPlay}
\newcommand{\soar}{SOAR}
\newcommand{\sgs}{SGS}

\title{CompassPlay: \\ Rewarding the Proposer for Where It Moves the Solver}

\author{%
Sophia Xiao Pu$^{1}$\quad Ximeng Sun$^{2}$\quad Jiang Liu$^{2}$\quad Jialian Wu$^{2}$\\
\textbf{Emad Barsoum$^{2}$\quad Zicheng Liu$^{2}$\quad William Yang Wang$^{1}$}\\[0.6ex]
{\normalfont $^{1}$University of California, Santa Barbara\quad $^{2}$Advanced Micro Devices}\\
{\normalfont \faEnvelope[regular]~\texttt{xiao\_pu@ucsb.edu}, \texttt{Ximeng.Sun@amd.com}\qquad
\faGlobe~\href{https://sophiapx.github.io/CompassPlay/}{Project Page}}
}

\begin{document}

\maketitle

\begin{abstract}
In self-play, a proposer generates verifiable tasks to train a solver.
Proposer rewards often depend on the solver's success rate, but equally difficult tasks can differ in their training value.
We introduce \cplay{}, a self-play method that rewards the proposer through gradient alignment.
The reward favors tasks whose solver loss gradients align with those of reference tasks representing the target capabilities.
It draws on a first-order approximation to learning progress and scores each eligible task without additional solver training.
Our experiments show gains in performance and training efficiency.
In coding self-play with Qwen2.5-Coder-7B, a small external reference set guides task generation.
\cplay{} improves average accuracy over AZR's difficulty reward by 1.5 percentage points on in-domain coding and 2.7 on out-of-domain mathematics.
In Lean~4 theorem proving, \cplay{} matches the difficulty baseline's 150-iteration cumulative coverage with 40\% fewer GPU-hours.
\end{abstract}

\begin{figure}[!b]
\centering
\includegraphics[width=\linewidth]{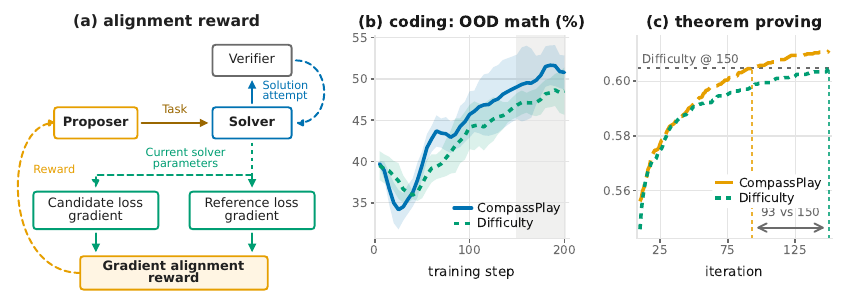}
\caption{\textbf{(a)} \cplay{} rewards alignment between the solver's loss gradients on verified candidate solutions and on reference solutions.
\textbf{(b)} Mean accuracy on five out-of-domain mathematics benchmarks during coding self-play.
Curves show the mean and $\pm 1$ s.d.\ across three seeds, averaged over three evaluation checkpoints.
\textbf{(c)} Lean self-play with novelty weighting: cumulative coverage of 3,323 target theorems from iteration 10, with one run per method.}
\label{fig:teaser}
\end{figure}

\section{Introduction}
\label{sec:intro}
Self-play turns task generation into part of reasoning-model training.
A proposer creates problems, a solver attempts them, and a verifier checks the answers \citep{zhao2025absolutezero,bailey2026sgs}.
This loop can expand training beyond a fixed collection of human-written problems.
Its value depends on the tasks that the proposer learns to generate.
The verifier checks whether a solution is correct, but does not establish how much training on generated tasks would help the solver.
The proposer therefore needs a reward that connects task generation to progress on the capabilities we want to improve.

A common strategy favors tasks near the solver's current ability \citep{zhao2025absolutezero,huang2025rzero}.
The solver's estimated success rate provides an inexpensive signal for finding such tasks.
This encourages a curriculum that adapts as the solver improves.
However, tasks with the same success rate can exercise different skills and contribute differently to target performance.
A reward based only on success rate cannot distinguish these contributions.

One way to estimate training value is to train on candidate tasks and measure the resulting improvement.
\soar{} follows this principle \citep{sundaram2026soar}.
It trains a copy of the solver on a group of generated problems and evaluates the copy on reference problems.
The accuracy gain becomes a shared reward for the group.
This measures learning progress directly, but requires additional training and evaluation for each group.
We seek a reward that estimates training value for individual tasks without these additional solver-training runs.

Gradient-based signals provide a natural proxy for this purpose. Prior work uses gradients to estimate training value for data selection \citep{xia2024less,yang2026gradalign} and influence-based feedback to guide synthetic-data generation \citep{li2025montessori,fan2026optimsyn}.
Building on these ideas, we introduce \textbf{\cplay}, a self-play method that rewards the proposer through gradient alignment.
For a proposed task with a verified solution, \cplay{} compares the solver's loss gradient on that solution with the gradient of a reference loss for the target capabilities (Figure~\ref{fig:teaser}a).
To first order, a small gradient step on the task lowers the reference loss in proportion to the inner product of the two gradients.
The inner product also grows with the task's gradient norm, which can rise without better alignment, so \cplay{} rewards the cosine of the two gradients instead (\S\ref{sec:method:reward}).
Both gradients are taken at the solver's current parameters, so the reference direction moves as the solver learns, and each eligible task receives its own score without additional solver training.

We evaluate \cplay{} on two different domains: coding and theorem proving. We construct the reference from the supervision available in each domain.
In coding self-play, the reference is a fixed external set of problems with gold solutions, which is not included in proposer prompts.
In our Lean setup, we use no external gold-proof reference set.
We instead use the solver's verified proofs of solved seed theorems as local references.
Here, the proposer also sees the seed statement.
We observe that optimizing alignment in this setting encourages renamed copies of the seed.
Weighting alignment by a novelty factor relative to the seed suppresses this behavior (\S\ref{sec:lean:novelty}).
Whether reference tasks appear in proposer prompts therefore matters for reward design.

Our contributions are as follows.
\begin{itemize}
\item We introduce \cplay{}, a proposer reward based on the cosine between candidate and reference gradients (\S\ref{sec:method}).
\item We show that a proposer prompted with the seed behind its reference learns to copy it, and that a novelty factor suppresses this copying (\S\ref{sec:lean:novelty}).
\item \cplay{} improves over AZR's difficulty reward \citep{zhao2025absolutezero} on all eight coding and mathematics benchmarks, by 1.5 points on coding and 2.7 on out-of-domain mathematics (\S\ref{sec:general:main}; Figure~\ref{fig:teaser}b). In Lean~4, it matches the difficulty baseline's coverage with 40\% fewer GPU-hours (\S\ref{sec:lean:cost}; Figure~\ref{fig:teaser}c).
\end{itemize}

\section{Method}
\label{sec:method}

\cplay{} changes the proposer reward within the self-play loop.
We first describe the self-play setting (\S\ref{sec:method:setup}), then present the reward design and reference choices (\S\ref{sec:method:reward}).

\subsection{Self-play with a proposer and a solver}
\label{sec:method:setup}

In a classic self-play setting, a proposer policy $\pi_\phi$ generates tasks for a solver policy $\pi_\theta$ to attempt (Figure~\ref{fig:teaser}a).
Their parameter vectors are $\phi$ and $\theta$, respectively.
The two policies may share weights, as in our coding experiments.
We denote a task by $x$ and a solver attempt by $y$.
Given a conditioning input $z$ (past tasks or a seed theorem), the proposer samples a candidate task
$\tilde{x} \sim \pi_\phi(\cdot \mid z)$.
The solver then samples $k$ attempts $y_1,\ldots,y_k \sim \pi_\theta(\cdot \mid \tilde{x})$.
A verifier $v$ (a Python executor or the Lean~4 compiler) assigns each attempt a binary correctness reward
$v(\tilde{x},y_j)\in\{0,1\}$.
The resulting solve rate is $s(\tilde{x})=\frac{1}{k}\sum_{j=1}^{k}v(\tilde{x},y_j)$.
The solver uses $v$ as its reward for policy-gradient reinforcement learning (RL).
The proposer maximizes the expectation of its scalar reward $r(\tilde{x})$:
\begin{equation}
  \max_\phi \; \mathbb{E}_{\tilde{x} \sim \pi_\phi(\cdot \mid z)} \big[ r(\tilde{x}) \big].
  \label{eq:method:objective}
\end{equation}

A common difficulty reward is $r(\tilde{x})=1-s(\tilde{x})$ for tasks with $0<s(\tilde{x})<1$,
optionally weighted by a judge score \citep{zhao2025absolutezero,bailey2026sgs}.

\subsection{\cplay{}: Gradient alignment as the proposer reward}
\label{sec:method:reward}

The difficulty term $1-s(\tilde{x})$ gives the same score to tasks with the same solve rate.
We propose to assess each task's training value through its effect on a reference loss.
\cplay{} estimates this effect using first-order gradients.

\paragraph{Candidate utility.}
To formalize learning benefit, let $\mathcal{L}_{\mathrm{ref}}(\theta)$ be a reference loss for the target problems.
For a held-out reference set, we use the average loss on its gold solutions.
When no such set is available, we use a local reference described below.
Let $g(\tilde{x})$ denote the gradient of the solver's loss on a candidate paired with a verified solution.
All gradients are taken with respect to the solver parameters $\theta$.
We define the candidate's \emph{utility} as the decrease in $\mathcal{L}_{\mathrm{ref}}$ under a hypothetical gradient step:
\begin{equation}
  U(\tilde{x}) \;=\; \mathcal{L}_{\mathrm{ref}}(\theta) \;-\; \mathcal{L}_{\mathrm{ref}}\big(\theta - \eta\, g(\tilde{x})\big),
  \qquad g(\tilde{x}) = \nabla_\theta\, \ell\big(\tilde{x}, y^\star(\tilde{x}); \theta\big),
  \label{eq:method:utility}
\end{equation}
where $\ell(x, y; \theta)$ is the solver's cross-entropy (CE) loss, averaged over solution tokens.
Here, $y^\star(\tilde{x})$ is a verified solution of the candidate, and $\eta$ is the step size.
This hypothetical loss-gradient step defines the scoring proxy.
Actual solver training uses the policy-gradient objective described above.

\paragraph{First-order approximation.}
We write $\gref$ for the gradient of the reference loss.
The symbols $\langle\cdot,\cdot\rangle$ and $\lVert\cdot\rVert$ denote the Euclidean inner product and norm, respectively.
For a small step size $\eta$, \eqref{eq:method:utility} expands to
\begin{equation}
  U(\tilde{x}) \;=\; \eta\, \big\langle g(\tilde{x}),\; \gref \big\rangle \;+\; O(\eta^2),
  \qquad \gref = \nabla_\theta \mathcal{L}_{\mathrm{ref}}(\theta).
  \label{eq:method:firstorder}
\end{equation}

This first-order term corresponds to the single-step influence estimate underlying TracIn \citep{pruthi2020tracin}.
GradAlign \citep{yang2026gradalign} uses a related gradient-alignment criterion to select RL training problems.
Evaluating this approximation requires no optimizer step or second-order derivatives.
Both gradients are taken at the solver's \emph{current} parameters, so the reference direction evolves as the solver learns.

\paragraph{From inner product to cosine.}

The inner product in \eqref{eq:method:firstorder} depends on both gradient norms and directional alignment.
For positive alignment, a larger candidate gradient norm can raise the score without improving that alignment.
\cplay{} therefore normalizes both gradients and uses their cosine similarity as its base proposer reward:
\begin{equation}
  r_{\cos}(\tilde{x}) \;=\; \cos\big(g(\tilde{x}),\, \gref\big)
  \;=\; \frac{\langle g(\tilde{x}), \gref\rangle}{\lVert g(\tilde{x})\rVert\,\lVert \gref\rVert}.
  \label{eq:method:cos}
\end{equation}

The resulting reward is bounded in $[-1,1]$ and invariant to positive rescaling of either gradient.
We compare cosine and inner-product rewards in our coding ablation (\S\ref{sec:general:ablations}; Appendix~\ref{app:cosdot}).

\paragraph{The reference gradient.}

\cplay{} constructs $\mathcal{L}_{\mathrm{ref}}$ from the reference data available in each domain.
In coding self-play, we use a held-out \emph{external} reference set $\mathcal{D}_{\mathrm{ref}} = \{(x_j, y_j^\star)\}$
of problems with gold solutions, which is not included in proposer prompts (\S\ref{sec:general} describes the set).
The reference loss is $\mathcal{L}_{\mathrm{ref}}(\theta) = \frac{1}{|\mathcal{D}_{\mathrm{ref}}|}\sum_j \ell(x_j, y_j^\star; \theta)$,
and $\gref$ is its gradient at the current $\theta$.

In our theorem-proving setup, we use no external gold-proof reference set.
The solver's own verified proofs provide the reference supervision.
Each candidate is generated from a seed statement that the solver has already proved.
We use the seed's verified proof as a \emph{local} reference, with
$\gref = \nabla_\theta \ell(x_{\mathrm{seed}}, y^\star_{\mathrm{seed}}; \theta)$.
The reward favors conjectures whose proof gradients align with this reference gradient.
The key distinction is whether reference tasks appear in proposer prompts.
In the local setting, the proposer sees the seed statement whose verified proof defines $\gref$.
We leave the dependence of $g(\tilde{x})$ and $\gref$ on $\theta$ implicit.
For local references, $\gref$ also depends on the candidate's seed.

\paragraph{The candidate gradient.}

\cplay{} computes $g(\tilde{x})$ from the solver's loss on a candidate paired with a \emph{verified} solution.
Unverified attempts are not used to compute this gradient.
In coding self-play, we use the oracle answer in the solver's answer format.
Depending on the task type, this answer is the executed output, the input supplied by the proposer, or the proposer's program.
In theorem proving, we average the loss gradients over up to $m$ verified proofs generated by the solver for each conjecture, with $m=4$ in our experiments.
We compute gradients with respect to all solver parameters and accumulate the cosine in single precision.

\paragraph{Which candidates are rewarded.}

\cplay{} applies the alignment reward to candidates solved in some but not all sampled attempts ($0 < s(\tilde{x}) < 1$).
In theorem proving, the base loop also requires $s(\tilde{x})$ to be at most the 70th percentile of solve rates in the batch.
Candidates outside these criteria receive no alignment reward.
We handle them as in the base loop: a fixed penalty in coding and exclusion from the proposer batch in theorem proving.
These rules govern proposer rewards; the solver retains its domain's training objective and task filters.
The proposer update retains the base loop's advantage estimator, reward normalization within each batch, and KL and clipping terms.

\paragraph{Lower computational cost.}

\cplay{} is computationally cheaper than measuring learning benefit through repeated solver training and evaluation.
It scores candidates individually using first-order gradients, without the additional training runs required by \soar{} \citep{sundaram2026soar} (Appendix~\ref{app:soar}).
The added work consists of forward/backward passes on verified candidate and reference solutions.
A shared reference gradient can be reused across candidates while the solver parameters remain fixed.
Scoring does not update model parameters, and temporary gradients are discarded afterward.
Relative to difficulty rewards, which reuse the solve-rate rollouts, the extra cost is the gradient computation.
We report measured costs and training efficiency in \S\ref{sec:general:oracle}, \S\ref{sec:lean:cost}, and Appendix~\ref{app:cost}.

\paragraph{Reward variants.}

In coding self-play, \cplay{} uses the cosine reward $r_{\cos}(\tilde{x})$ from \eqref{eq:method:cos}.
In theorem proving, alignment with a local reference can reward copies of the seed (\S\ref{sec:lean:collapse}).
We therefore multiply the cosine reward by a novelty factor.
Let $J(\tilde{x},x_{\mathrm{seed}})\in[0,1]$ denote the token-Jaccard similarity between the candidate
statement $\tilde{x}$ and its seed $x_{\mathrm{seed}}$.
The resulting reward is
\begin{equation}
  r_{\cos \times \mathrm{nov}}(\tilde{x})
  = r_{\cos}(\tilde{x})\,
    \bigl(1-J(\tilde{x},x_{\mathrm{seed}})\bigr).
  \label{eq:method:cosnov}
\end{equation}
In the experiments, \cplay{} (\cosm) uses $r_{\cos}$, while \cplay{} (\cosnov) uses $r_{\cos \times \mathrm{nov}}$.
Appendix~\ref{app:instantiation} summarizes both instantiations and gives the self-play loop in Algorithm~\ref{alg:loop}, using the same similarity function $J$.

\section{General reasoning with coding self-play}
\label{sec:general}

\paragraph{Setup.}
We evaluate \cplay{} in the coding self-play setting of AZR \citep{zhao2025absolutezero}.
The proposer and solver share a single Qwen2.5-Coder-7B model \citep{hui2024qwen25coder}.
Tasks are built from Python programs and verified by execution.
Task types, the solver objective, and the training configuration follow AZR.
We compare three configurations:
\begingroup
\setlength{\leftmargini}{0pt}
\begin{itemize}
  \item \textbf{\solveonly}: only the solver objective is optimized.
    Proposals still come from the shared weights, which change as the solver learns.
  \item \textbf{AZR (\difficulty)}: the difficulty reward of AZR, $1-s$ for $0<s<1$.
  \item \textbf{\cplay{} (\cosm)}: the cosine gradient-alignment reward in \eqref{eq:method:cos}.
\end{itemize}
\endgroup
AZR and \cplay{} differ only in the proposer reward.
For \cplay{}, $\gref$ is the cross-entropy gradient of gold solutions to 32 problems from the MATH training split.
This reference set is fixed throughout training, disjoint from evaluation, and not included in proposer prompts.
Mathematics problems are used only for the reference gradient; solver-training tasks remain coding problems.

Each configuration runs for 200 steps.
We evaluate with greedy decoding every 5 steps on three coding benchmarks:
LiveCodeBench \citep{jain2024livecodebench}, CRUXEval-I, and CRUXEval-O \citep{gu2024cruxeval}.
We also evaluate on five mathematics benchmarks:
MATH500 \citep{lightman2023verify}, AMC23, GSM8K \citep{cobbe2021gsm8k},
Minerva Math \citep{lewkowycz2022minerva}, and OlympiadBench \citep{he2024olympiadbench}.
Mathematics is out of domain with respect to the generated training tasks.
\cplay{} and AZR use the same three random seeds; \solveonly{} has one run.
Appendix~\ref{app:details} gives benchmark and evaluation details.

\subsection{Main results}
\label{sec:general:main}

Tables~\ref{tab:general:indomain} and~\ref{tab:general:ood} average the 11 evaluations at steps 150--200 within each run, then report the mean and sample standard deviation across runs.
The Mean column weights benchmarks equally.

\begin{table}[t]
\centering
\caption{In-domain coding accuracy (\%). Trained methods are averaged over steps 150--200.
\cplay{} and AZR report mean $\pm$ sample s.d.\ across three seeds; \solveonly{} has one run.
Initial model is Qwen2.5-Coder-7B before self-play.
Mean averages benchmarks equally. Gain over AZR is the \cplay{} improvement in percentage points (pp).}
\label{tab:general:indomain}
\small
\begin{tabular}{lcccc}
\toprule
Method & LiveCodeBench & CRUXEval-I & CRUXEval-O & Mean \\
\midrule
Initial model & 40.02 & 48.98 & 50.89 & 46.63 \\
\midrule
\solveonly{} & 46.78 & 58.05 & 60.20 & 55.01 \\
AZR & 49.87 $\pm$ 1.62 & 59.63 $\pm$ 0.12 & 61.23 $\pm$ 0.47 & 56.91 $\pm$ 0.72 \\
\cplay{} (\cosm) & 52.60 $\pm$ 0.17 & 60.08 $\pm$ 0.34 & 62.56 $\pm$ 0.44 & \textbf{58.41 $\pm$ 0.18} \\
\midrule
Gain over AZR (pp) & +2.73 & +0.45 & +1.33 & \textbf{+1.50} \\
\bottomrule
\end{tabular}
\end{table}

\begin{table}[t]
\centering
\caption{Out-of-domain mathematics accuracy (\%), using the same averaging as Table~\ref{tab:general:indomain}.
Initial model reports accuracy before self-play.
Gain over AZR is the \cplay{} improvement in percentage points (pp).}
\label{tab:general:ood}
\small
\setlength{\tabcolsep}{4pt}
\resizebox{\linewidth}{!}{%
\begin{tabular}{lcccccc}
\toprule
Method & MATH500 & AMC23 & GSM8K & Minerva & OlympiadBench & Mean \\
\midrule
Initial model & 53.40 & 45.00 & 67.00 & 18.00 & 19.30 & 40.54 \\
\midrule
\solveonly{} & 66.11 & 37.73 & 78.25 & 26.84 & 31.22 & 48.03 \\
AZR & 60.73 $\pm$ 2.63 & 43.94 $\pm$ 0.92 & 76.98 $\pm$ 1.83 & 26.54 $\pm$ 1.40 & 30.40 $\pm$ 2.04 & 47.72 $\pm$ 1.07 \\
\cplay{} (\cosm) & 63.34 $\pm$ 3.77 & 47.12 $\pm$ 6.56 & 78.72 $\pm$ 1.95 & 29.14 $\pm$ 2.34 & 33.53 $\pm$ 1.58 & \textbf{50.37 $\pm$ 2.87} \\
\midrule
Gain over AZR (pp) & +2.61 & +3.18 & +1.74 & +2.60 & +3.14 & \textbf{+2.65} \\
\bottomrule
\end{tabular}}
\end{table}

\begin{figure}[t]
\centering
\includegraphics[width=\linewidth]{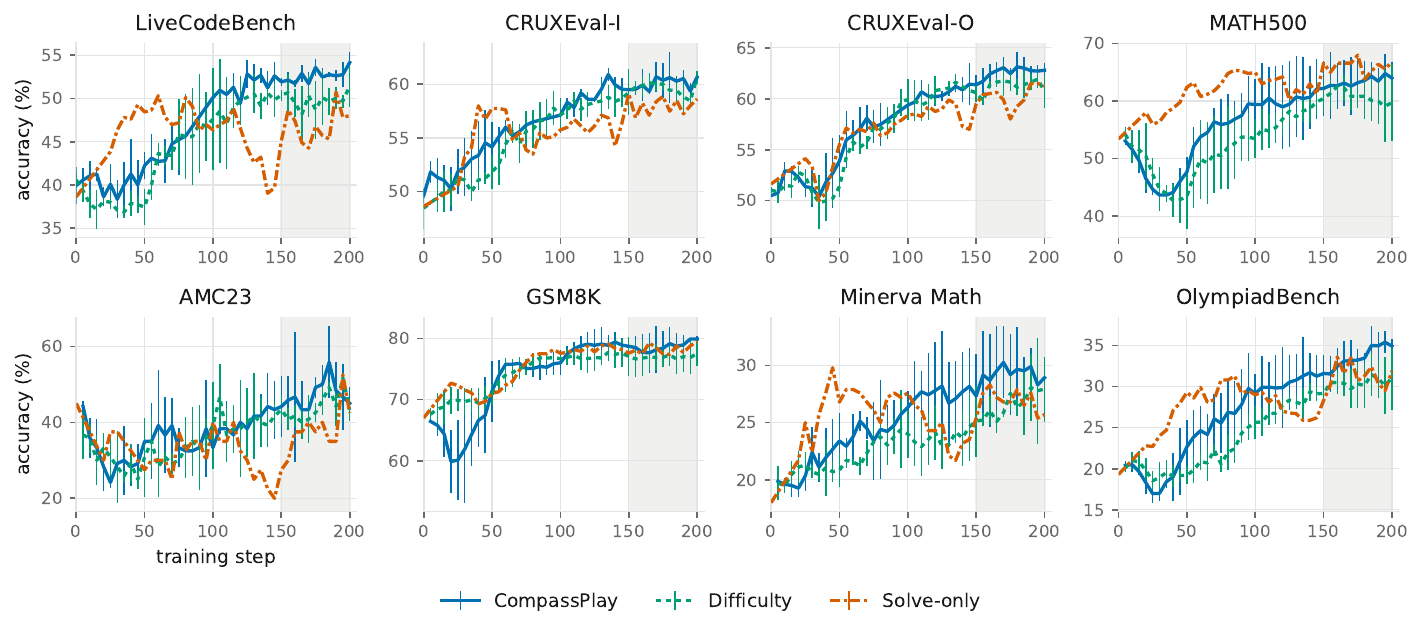}
\caption{Accuracy over training on three coding and five mathematics benchmarks.
The figure labels AZR as \difficulty{}.
These two methods report means over three seeds with $\pm 1$ sample s.d.\ error bars; \solveonly{} is a single run.
The shaded band marks steps 150--200, the averaging window in Tables~\ref{tab:general:indomain} and~\ref{tab:general:ood}.}
\label{fig:general:curves}
\end{figure}

\cplay{} improves mean accuracy over AZR on all eight benchmarks.
The average gains are 1.50 percentage points on coding and 2.65 on mathematics.
Against the single \solveonly{} run, \cplay{} gains 3.40 and 2.34 percentage points on coding and mathematics, respectively.
Figure~\ref{fig:general:curves} shows a sustained lead for \cplay{} over AZR late in training on LiveCodeBench.

\subsection{Comparison with SOAR}
\label{sec:general:oracle}

\cplay{} estimates each candidate's training value from gradient alignment.
\soar{} \citep{sundaram2026soar} instead rewards candidate groups by the reference-accuracy gain after training a solver copy on each group.
We compare the downstream accuracy and computational cost of these approaches.
No public implementation was available, so we reimplemented \soar{} on our training stack.
It shares the base model, proposer, and outer optimizer with the other methods.
Appendix~\ref{app:soar} describes the inner loop, reference set, and deviations from the original recipe.
We ran this baseline once for 20 outer steps.

\begin{figure}[t]
\centering
\includegraphics[width=\linewidth]{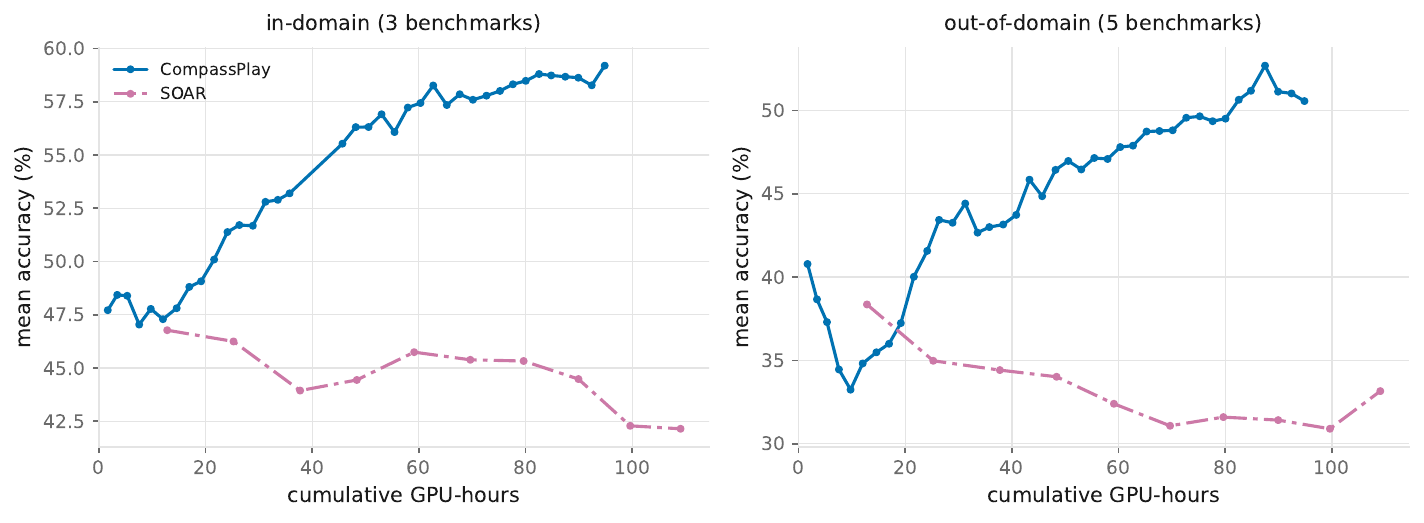}
\caption{Mean coding and mathematics accuracy against cumulative GPU-hours.
\cplay{} (\cosm) reports mean accuracy and cost across three seeds; \soar{} is our single-run reimplementation.
Each point is an evaluation checkpoint.
GPU-hours are training wall-clock time multiplied by the number of GPUs, excluding benchmark evaluation.
Appendix~\ref{app:cost} gives the comparison with AZR at matched GPU-hours.}
\label{fig:general:tradeoff}
\end{figure}

Figure~\ref{fig:general:tradeoff} shows that \cplay{} achieves higher final accuracy at lower total training cost than our \soar{} reimplementation.
A 200-step \cplay{} run uses 95 GPU-hours on average across three seeds, while the single 20-step \soar{} run uses 109 GPU-hours.

The cost difference reflects how the rewards are computed.
In our \soar{} reimplementation, each outer step requires 16 inner RL runs; the observed rewards take 2--4 distinct values shared across 256 candidates.
\cplay{} instead scores each eligible candidate using candidate and reference gradients, without training a solver copy.
Appendix~\ref{app:cost} reports the per-phase costs and the comparison with AZR at matched GPU-hours.

\subsection{Ablations}
\label{sec:general:ablations}

We test the two design choices in \S\ref{sec:method:reward}: which gradients represent the candidate and the reference, and how the two gradients are compared. Each configuration is run once.

\paragraph{CE and REINFORCE gradients.}
The solver is trained with policy-gradient RL, whereas \cplay{} computes both gradients from the CE loss on verified or gold solutions (\S\ref{sec:method:reward}).
We therefore test whether gradients of the RL objective give a better reward.
We replace CE gradients with REINFORCE estimates of the gradient of the expected verifier reward on the reference side alone or on both sides \citep{williams1992reinforce}.
The CE configuration outperforms both variants at the CE run's final compute budget.
With a shared seed, the CE run uses 93 GPU-hours for 200 training steps, versus 161 with REINFORCE gradients on both sides (Appendix~\ref{app:refgrad}).

\paragraph{Cosine and dot-product rewards.}
The first-order utility in \eqref{eq:method:firstorder} is the inner product, which also grows with the candidate gradient norm.
We test whether normalizing it to the cosine improves training.
Cosine achieves higher mean accuracy than the dot product over the final quarter of training: 58.31\% vs.\ 55.99\% on coding and 48.13\% vs.\ 45.16\% on mathematics.
Appendix~\ref{app:cosdot} gives the settings and learning curves.

\FloatBarrier
\section{Theorem proving with Lean self-play}
\label{sec:lean}

\paragraph{Setup.}
We evaluate \cplay{} on the 3,323 target theorems of SGS \citep{bailey2026sgs} using separate DeepSeek-Prover-V2-7B models for the proposer and solver \citep{ren2025deepseekproverv2}.
We use full-parameter updates and check proofs with Lean~4 \citep{moura2021lean4} against mathlib4 \citep{mathlib2020}.
Our implementation adapts the SGS training framework to condition conjectures on solved target theorems (seeds); the original SGS recipe conditions on unsolved targets.
For each candidate, $\gref$ is computed from its seed's verified proof at the current solver parameters.
We use no external gold-proof reference set.

In each iteration, the solver attempts every unsolved target theorem.
We report cumulative coverage on this target set: the fraction of target theorems solved at least once.
Each method is run once.

\paragraph{Methods.}
We compare proposer-training strategies within the same adapted setup:
\begingroup
\setlength{\leftmargini}{0pt}
\begin{itemize}
  \item \textbf{\solveonly}: the proposer is frozen, and only the solver is trained.
  \item \textbf{\sgs{} reward}: the SGS guide score times the difficulty term $1-s$.
  \item \textbf{\difficulty}: the solve-rate reward $1-s$.
  \item \textbf{\cplay{} (\cosm)}: the cosine reward in \eqref{eq:method:cos}, $\cossim$, with the local reference.
  \item \textbf{\cplay{} (\cosnov)}: alignment weighted by novelty relative to the seed (\S\ref{sec:lean:novelty}).
\end{itemize}
\endgroup

\subsection{Copying with a local reference}
\label{sec:lean:collapse}

\begin{figure}[tbp]
\centering
\includegraphics[width=0.65\linewidth]{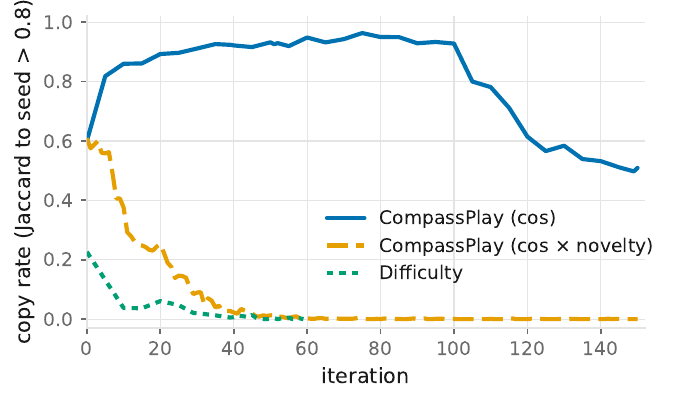}
\caption{Copy rate: the fraction of eligible conjectures with token-Jaccard similarity to the seed above 0.8.
The \cosm{} and \cosnov{} curves denote the two \cplay{} variants.}
\label{fig:lean:copy}
\end{figure}

The proposer sees the seed whose proof defines its local reference.
Repeating that seed can therefore offer a shortcut to high alignment without introducing a new task.
We measure copying by the fraction of eligible conjectures with token-Jaccard similarity above 0.8 to their seed.
In Figure~\ref{fig:lean:copy}, the copy rate of \cplay{} (\cosm) reaches 96\% at iteration 75, while \difficulty{} falls below 1\% by iteration 40.
A separate name-normalization audit confirms that many conjectures are pure renamings of their seeds (Appendix~\ref{app:rename}).

\subsection{Novelty-weighted alignment}
\label{sec:lean:novelty}

To reduce copying in this setting, we multiply the \cosm{} reward by novelty relative to the seed, giving \cplay{} (\cosnov) in \eqref{eq:method:cosnov}.
The novelty factor is zero at token-Jaccard similarity 1 and shrinks the alignment score's magnitude as overlap increases.
We retain eligible candidates in the proposer batch and apply the base loop's reward normalization.

With \cplay{} (\cosnov), the copy rate stays below 0.5\% over iterations 60--150 (Figure~\ref{fig:lean:copy}).
At the audited checkpoints, the pure-rename rates are 1.0\% for \cosnov{} and 77.5\% for \cosm{} (Appendix~\ref{app:rename}).

\subsection{Coverage and computational cost}
\label{sec:lean:cost}

\begin{figure}[tbp]
\centering
\subfloat[\label{fig:lean:coverage}]{%
  \includegraphics[width=0.48\linewidth]{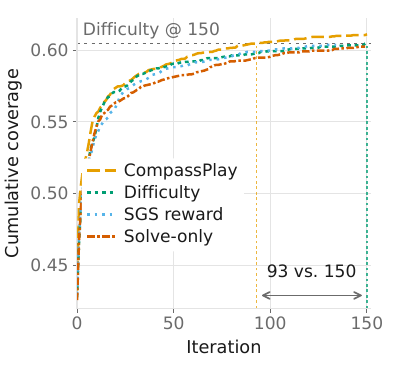}%
}\hfill
\subfloat[\label{fig:lean:tradeoff}]{%
  \includegraphics[width=0.48\linewidth]{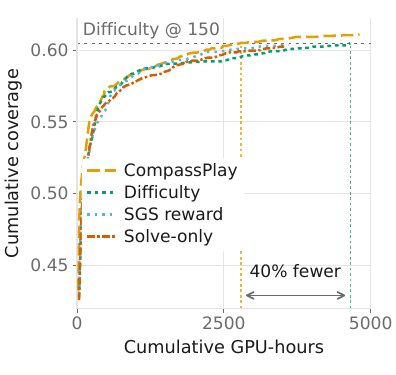}%
}
\caption{Cumulative coverage of \cplay{} (\cosnov) and baselines at available checkpoints through iteration 150.
The dotted line marks \difficulty{}'s coverage at iteration 150.
GPU-hours are computed as training wall-clock time times 8 GPUs.}
\label{fig:lean:results}
\end{figure}

\cplay{} (\cosnov) reaches \difficulty{}'s 150-iteration cumulative coverage in 93 iterations (Figure~\ref{fig:lean:coverage}).
Reaching that coverage costs 2,787 versus 4,652 GPU-hours, a 40\% reduction (Figure~\ref{fig:lean:tradeoff}).
At iteration 150, the two methods have solved 2,031 vs.\ 2,010 targets.

\subsection{Ablation}
\label{sec:lean:ablation}

We remove each factor of the \cosnov{} reward in turn (Table~\ref{tab:lean:ablation}).
The novelty-only variant uses the novelty factor $1-J(\tilde{x},x_{\mathrm{seed}})$ alone as the proposer reward.
\cplay{} (\cosnov) and the novelty-only variant both have zero copy rate at iteration 115, yet their mean gains over \difficulty{} across iterations 15--115 are +13.6 and $-7.9$ solved targets, respectively.
The combined reward therefore achieves better coverage than novelty alone in these runs.

\begin{table}[tbp]
\centering
\caption{Ablation of \cplay{} (\cosnov).
Copy rate is measured at iteration 115.
Gain is the mean number of additional targets solved over \difficulty{}, using available paired checkpoints in iterations 15--115.}
\label{tab:lean:ablation}
\small
\setlength{\tabcolsep}{4pt}
\begin{tabular}{lcccc}
\toprule
Method & Alignment & Novelty & Copy rate $\downarrow$ & Gain (targets) $\uparrow$ \\
\midrule
\cplay{} (\cosm) & $\checkmark$ & $\times$ & 0.712 & +3.1 \\
\textsc{novelty-only} & $\times$ & $\checkmark$ & 0.000 & $-7.9$ \\
\cplay{} (\cosnov) & $\checkmark$ & $\checkmark$ & \textbf{0.000} & \textbf{+13.6} \\
\bottomrule
\end{tabular}
\end{table}

\section{Related work}
\label{sec:related}

\paragraph{Proposer rewards in self-play.}
Many self-play methods use solver feedback to guide task generation \citep{zhao2025absolutezero,bailey2026sgs,dong2025stp,poesia2024minimo}.
Some favor tasks near the solver's current ability using estimates of success or uncertainty \citep{huang2025rzero,liu2025spice}.
Others reward tasks that cause solver failure \citep{zhang2025dualplay,kuba2025lsp}.
SGS supplements difficulty with an LLM guide score \citep{bailey2026sgs}.
GASP uses hard goalpost problems to guide generation \citep{jana2026gasp}, while \citet{pu2026survive} study data gating and reward grounding.
\soar{} rewards task groups by the gain in reference accuracy after training a solver copy on them \citep{sundaram2026soar}.
\cplay{} uses gradient alignment as a proxy for learning progress.
Each eligible candidate receives a score without these additional solver-training runs.

\paragraph{Learnability signals.}
Automatic curricula prioritize tasks using intermediate success rates \citep{florensa2018goalgan}, learning progress \citep{matiisen2017tscl}, or regret \citep{dennis2020paired,parkerholder2022accel}.
In LLM reinforcement learning, SEC uses a learnability-based curriculum \citep{chen2025sec}, and LearnAlign uses gradient alignment for data selection \citep{li2025learnalign}.
\cplay{} uses a related alignment signal to train a proposer to generate tasks.

\paragraph{Influence-based data selection and generation.}
TracIn estimates data influence by accumulating gradient inner products across training checkpoints \citep{pruthi2020tracin}.
LESS selects instruction data using gradient similarity to target examples \citep{xia2024less}.
MATES learns an influence model for pretraining data selection \citep{yu2024mates}.
GradAlign uses alignment between candidate and validation gradients for RL data selection \citep{yang2026gradalign}.
Influence feedback can also guide generation.
Montessori-Instruct trains a teacher using changes in reference loss for individual training examples and updates the student between rounds \citep{li2025montessori}.
OptimSyn uses optimizer-aware gradient influence to reward a rubric generator \citep{fan2026optimsyn}.
\cplay{} uses gradient alignment in self-play with executable or formal verification.
Both proposer and solver learn throughout training, and the reward uses gradients at the current solver parameters.
In Lean, we observe substantial copying when the proposer sees the seed defining its local reference.
Novelty weighting reduces this behavior (\S\ref{sec:lean:novelty}).
In our coding setup, reference tasks are not included in proposer prompts and no novelty factor is used.

\section{Conclusion}
\label{sec:conclusion}

\cplay{} rewards proposed tasks through gradient alignment with a reference objective.
It scores each eligible candidate without training a separate solver copy.
In coding self-play, \cplay{} uses an external reference and improves average accuracy over AZR on both coding and mathematics.
In Lean, the solver's verified seed proofs provide local references, but the proposer can learn to copy the seeds it sees.
With novelty weighting, \cplay{} reduces copying and matches the difficulty baseline's 150-iteration cumulative coverage with 40\% fewer GPU-hours.
These results support gradient alignment as a practical proposer reward and highlight copying as a concern when reference tasks are also used to prompt the proposer.

\clearpage

\bibliography{references}
\bibliographystyle{arxiv}

\clearpage

\appendix
\section{The SOAR baseline}
\label{app:soar}

\soar{} \citep{sundaram2026soar} is an asymmetric self-play method whose proposer reward measures solver learning progress.
Our reimplementation samples $g = 4$ groups of $n = 64$ candidate problems per outer step.
It trains a solver copy on each group for $r = 4$ repeated inner RL runs of 10 steps,
evaluates the copy on a held-out reference set of hard problems, and rolls it back.
Every candidate in a group receives the group's accuracy gain:
\begin{equation*}
  R(X_k) = \mathrm{Acc}(\theta'_k, Q_R) - \mathrm{Acc}(\theta, Q_R).
\end{equation*}
Here $X_k$ is the $k$-th group, $Q_R$ is the reference set, $\theta$ denotes the current solver parameters,
and $\mathrm{Acc}(\theta'_k, Q_R)$ is the reference accuracy after training on $X_k$, averaged over the $r$ repeated runs.

Each outer step of our reimplementation requires $g \times r = 16$ inner RL runs and 17 reference-set evaluations.
It produces at most $g$ distinct reward values per outer step (2--4 observed in our run),
with each group reward shared by all 64 candidates in that group.
\cplay{} (\cosm) assigns a score to each eligible candidate individually (Figure~\ref{fig:general:tradeoff}).

No public implementation of \soar{} was available, so we reimplemented it from the paper on our training stack,
keeping the base model, the proposer and the outer optimizer the same as for the other methods. Following the
paper, the reference set is drawn from held-out problems the base model cannot solve (a pool of ${\sim}370$
MATH-train problems with 0/32 successes, 92 resampled every outer step); the reference pool of \cosm{} is drawn
from the same held-out MATH-train split but without the difficulty filter. The recipe could not be followed
exactly on a single-model self-play stack; the five deviations are: the inner loop uses the outer REINFORCE++
\citep{hu2025reinforcepp} objective instead of RLOO \citep{ahmadian2024rloo} because the stack samples one rollout per problem; one problem type per inner step;
the group reward replaces the whole proposer reward rather than only its difficulty term; promotion updates
weights but not the teacher on that step, and not the optimizer state. \soar{} was run once, for 20 outer
steps. Its outer step contains 16 inner RL runs and 17 reference-set evaluations (measured on a smoke run:
about 10.8~s per inner step and 20~s per evaluation); its 20 outer steps cost 109 GPU-hours in total.

\section{Per-step cost breakdown}
\label{app:cost}

GPU-hours are training wall-clock hours multiplied by the number of GPUs.
Coding and Lean runs use different GPUs (Appendix~\ref{app:details}), so GPU-hours are comparable within a domain but not across domains.
Periodic benchmark evaluations are excluded; \soar{}'s internal reference evaluations are included.
To locate the overhead of \cplay{} (\cosm), Table~\ref{tab:cost:phases} compares it with AZR (\difficulty).
The configurations share the training loop and differ in the proposer reward.
Per-phase timings come from the training framework's timers on one run of each method (8 GPUs, mean over the run).
A step is split into three phases: \emph{proposer reward}, the solve-rate estimation rollouts and reward computation,
including gradient alignment for \cosm{}; \emph{optimizer update}, the forward/backward pass and parameter update
on the training batch; and \emph{remainder}, candidate and solution generation, log-probability passes,
weight synchronization to the sampler, and bookkeeping.
Cells are seconds per step, with the phase's share of the step in parentheses.

\begin{table}[h]
\centering
\caption{Per-phase wall-clock of one training step (seconds; share of the step in parentheses), one run of each
method on 8 GPUs, mean over the run.}
\label{tab:cost:phases}
\small
\begin{tabular}{lcccc}
\toprule
reward & proposer reward (s) & optimizer update (s) & remainder (s) & step total (s) \\
\midrule
\difficulty    & 72.1 (44\%)  & 13.7 (8\%) & 77.1 (47\%) & 162.9 \\
\cosm{} (ours) & 128.8 (58\%) & 13.4 (6\%) & 80.2 (36\%) & 222.4 \\
\bottomrule
\end{tabular}
\end{table}

Most of the \cosm{} overhead is in the proposer-reward phase (+56.7~s of about 60~s per step),
which includes candidate and reference gradient computation.
Optimizer-update time remains similar (13.7 vs.\ 13.4~s).

The optimizer update accounts for under 10\% of a step.
\solveonly{} has a similar total training cost to \difficulty{} (70 vs.\ 71 GPU-hours per 200 steps).
It still samples 192 proposals per step from the shared weights, has the solver attempt them,
and runs the same solve-rate estimation.
It trains about a third of the tokens, with the PPO \citep{schulman2017ppo} mini-batch halved automatically,
but optimizer updates account for only a small fraction of the total cost.
\difficulty{} computes its reward from the existing solve-rate rollouts without additional gradient computation.

\paragraph{Matched-GPU-hours readout for the methods whose per-step cost differs.}
Table~\ref{tab:cost:matched} gives, for each budget, the last evaluation point with cumulative cost within the
budget.

\begin{table}[h]
\centering
\caption{Mean accuracy (\%) at matched GPU-hours.
For \cplay{} (\cosm) and AZR (\difficulty), accuracy and cumulative cost are averaged across three seeds
at each evaluation step; \soar{} is our single-run reimplementation.
Each budget uses the last step whose mean cumulative cost is within budget.
The endpoint row gives final-checkpoint accuracy and total GPU-hours in parentheses;
these are distinct from the final-quarter averages in Tables~\ref{tab:general:indomain} and~\ref{tab:general:ood}.}
\label{tab:cost:matched}
\small
\setlength{\tabcolsep}{3.5pt}
\resizebox{\linewidth}{!}{%
\begin{tabular}{lcccccc}
\toprule
& \multicolumn{3}{c}{in-domain} & \multicolumn{3}{c}{OOD} \\
\cmidrule(lr){2-4} \cmidrule(lr){5-7}
budget (GPU-h) & \cosm & \difficulty & \soar & \cosm & \difficulty & \soar \\
\midrule
20  & 49.07 & 48.94 & 46.78 & 37.24 & 39.33 & 38.36 \\
40  & 53.20 & 54.98 & 43.94 & 43.15 & 43.65 & 34.42 \\
60  & 57.23 & 57.38 & 45.75 & 47.09 & 46.85 & 32.40 \\
69  & 57.86 & 56.90 & 45.75 & 48.77 & 48.10 & 32.40 \\
endpoint & 59.20 (95) & 57.60 (71) & 42.15 (109) & 50.56 (95) & 47.93 (71) & 33.16 (109) \\
\bottomrule
\end{tabular}}
\end{table}

\section{Ablations on the coding task}
\label{app:ablations}

The two ablations summarized in \S\ref{sec:general:ablations}, in full.

\subsection{The reference gradient}
\label{app:refgrad}

\paragraph{Question.}

The reward $\cossim$ needs a reference gradient $\gref$ that stands for ``progress on the held-out problems''.
The main method uses the cross-entropy (CE) loss gradient on gold solutions.
This is cheap and low-variance, but the solver itself is trained with RL.
Does a REINFORCE loss-gradient estimate give a better signal, and at what cost?

\paragraph{Methods.}

We compare three gradient configurations.
The reference-sampling settings also differ; Appendix~\ref{app:details} gives those used by the REINFORCE variants.

\begin{figure}[t]
\centering
\includegraphics[width=\linewidth]{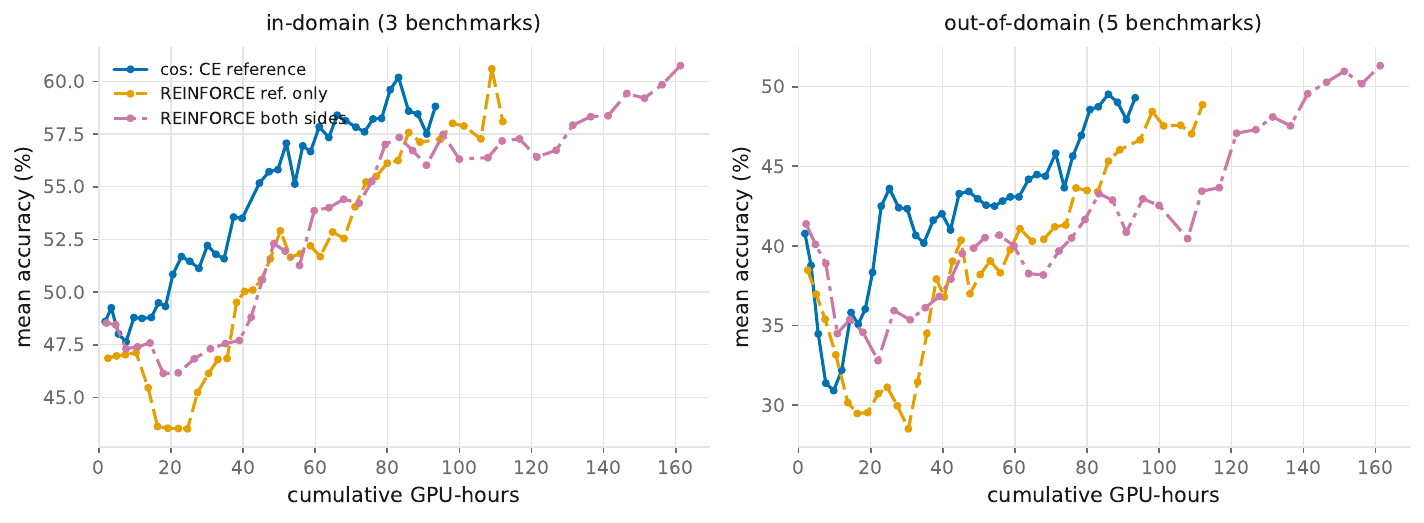}
\caption{Reference-gradient ablation: CE reference (the main method) against REINFORCE reference gradients,
accuracy against cumulative GPU-hours. Single runs with a shared seed.}
\label{fig:general:celoss}
\end{figure}

\begin{itemize}
  \item \textbf{\cosm{} (CE reference)}: both sides use CE loss gradients of gold solutions.
    The CE loss is averaged over solution tokens. This is the main method.
  \item \textbf{REINFORCE reference only}: we use a REINFORCE
    \citep{williams1992reinforce} estimate of $\nabla_\theta\,\mathbb{E}[v(x,y)]$ on held-out problems
    (Appendix~\ref{app:details}), where $v(x,y)$ is the solver's verifier reward.
    For alignment, the reference gradient is the negative of this estimate.
    The candidate side keeps the CE loss gradient.
  \item \textbf{REINFORCE both sides}: both sides use the negative of their respective REINFORCE
    reward-gradient estimates, so both are loss gradients.
\end{itemize}
Each configuration is a single run with the same seed. Figure~\ref{fig:general:celoss} has the form of
Figure~\ref{fig:general:tradeoff}.

\paragraph{Results.}

At the CE run's final compute budget (93 GPU-hours), \cosm{} has higher in-domain and OOD accuracy than both REINFORCE configurations.
The configuration with REINFORCE gradients on both sides uses 161 GPU-hours for 200 steps, compared with 93 for \cosm{} with the same seed, a factor of $1.7\times$.
The reference-only variant combines a REINFORCE loss-gradient estimate on the reference with CE loss gradients on candidates.
Its OOD accuracy drops below 30\% in the first 30 steps and ends below \cosm{}.

\subsection{Cosine vs dot product}
\label{app:cosdot}

\paragraph{Question.}

The dot product $\langle \gsynth, \gref \rangle$ in the first-order approximation depends on both gradient norms
and directional alignment. For positive alignment, a larger candidate gradient norm $\|\gsynth\|$ can raise
the score without changing the gradient direction. Cosine is invariant to positive rescaling of either gradient.
Does this normalization improve training?

\paragraph{Methods.}
Two runs that share the seed and differ in one flag, cosine vs dot, both with REINFORCE gradients on both sides
(the REINFORCE-both-sides configuration of Appendix~\ref{app:refgrad} is the cosine member).
Figure~\ref{fig:general:cosdot} has the form of Figure~\ref{fig:general:tradeoff}.

\begin{figure}[t]
\centering
\includegraphics[width=\linewidth]{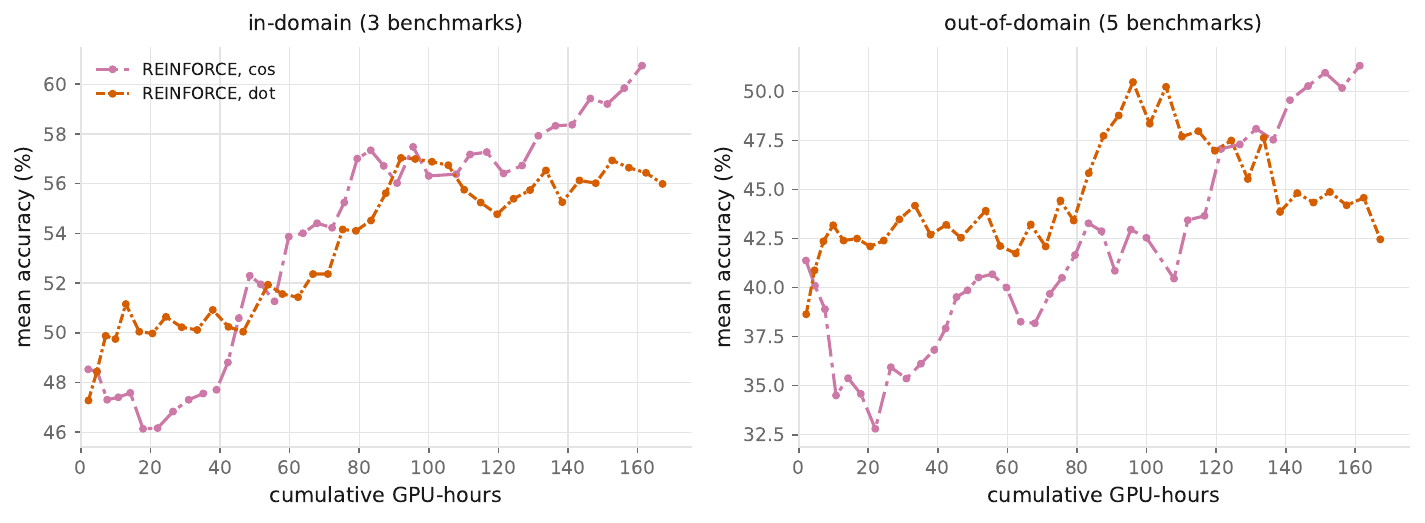}
\caption{Cosine against dot product as the alignment score, accuracy against cumulative GPU-hours, with
REINFORCE gradients on both sides (Appendix~\ref{app:refgrad}). The two runs share the seed and differ
in one flag.}
\label{fig:general:cosdot}
\end{figure}

\paragraph{Results.}
Dot leads OOD early and mid-run, the curves cross around step 160, and \textbf{cosine pulls away from there}:
over the final quarter of the run it is ahead by 2.3 points in-domain and 3.0 OOD. Dot peaks mid-run on both
axes and then declines, whereas cosine peaks at the last step.
For cosine versus dot product, mean accuracy over steps 150--200 is 58.31\% versus 55.99\% in-domain and 48.13\% versus 45.16\% OOD.
This is the averaging window of Tables~\ref{tab:general:indomain} and~\ref{tab:general:ood}, applied here unchanged.

\section{Pure-rename audit}
\label{app:rename}

For each method, the eligible conjectures of one iteration are compared with
their seed after normalizing theorem names, bound variable names and whitespace. \emph{pure-rename} is exact
equality after normalization; \emph{near-copy} is a token-Jaccard of at least 0.9 after normalization;
\emph{Jaccard $> 0.8$} is the raw-token criterion used in Figure~\ref{fig:lean:copy}. All values in percent.

The raw-token rates here are recomputed for the audit.
The \cosm{} and \cosnov{} curves use online measurements with slightly different tokenization,
so the audited and plotted Jaccard rates at the same iteration can differ.

\begin{table}[h]
\centering
\caption{Pure-rename and token-overlap rates at each method's audited iteration (percent).}
\label{tab:rename}
\small
\begin{tabular}{lcccc}
\toprule
Method & Iteration & Pure-rename \% & Near-copy \% & Jaccard $> 0.8$ \% \\
\midrule
\cplay{} (\cosm)   & 39 & \textbf{77.50} & 92.20 & 91.06 \\
\cplay{} (\cosnov) & 40 & \textbf{1.01}  & 4.48  & 1.46 \\
\difficulty & 39 & \textbf{0.27}  & 2.45  & 0.54 \\
\bottomrule
\end{tabular}
\end{table}

\section{Experimental details}
\label{app:details}

\paragraph{Hardware.}
Coding runs use 8 AMD MI355X GPUs; Lean runs use 8 AMD MI250X GPUs. GPU-hours are therefore comparable
within a domain but not across domains.

\paragraph{Benchmarks (Section~\ref{sec:general}).}
In-domain: LiveCodeBench \citep{jain2024livecodebench}, competitive-programming problems collected after the
base model's training cutoff, scored by hidden tests; CRUXEval-I and CRUXEval-O \citep{gu2024cruxeval},
predicting the input, respectively the output, of a short Python function, the two directions of the
code-reasoning task that self-play trains on. Out-of-domain: MATH500 \citep{lightman2023verify}, a 500-problem
subset of the MATH competition benchmark \citep{hendrycks2021math}; AMC23, the 40 problems of the 2023
American Mathematics Competitions; GSM8K \citep{cobbe2021gsm8k}, grade-school word problems; Minerva Math
\citep{lewkowycz2022minerva}, undergraduate STEM problems; and OlympiadBench \citep{he2024olympiadbench},
olympiad-level mathematics. Evaluation is greedy (temperature 0, $k = 1$).

\paragraph{Evaluation window.}

Tables~\ref{tab:general:indomain} and~\ref{tab:general:ood} average over the final quarter of training,
steps 150--200, to reduce sensitivity to any single checkpoint.
Greedy evaluations fluctuate from one checkpoint to the next (Figure~\ref{fig:general:curves}).

\paragraph{REINFORCE reference gradient (Appendix~\ref{app:ablations}).}

The REINFORCE estimator targets $\nabla_\theta\,\mathbb{E}[v(x,y)]$.
The expectation averages over reference problems $x$ and solver responses $y\sim\pi_\theta(\cdot\mid x)$;
$v(x,y)$ is the binary correctness reward defined in \S\ref{sec:method:setup}.
For alignment, we use the corresponding loss gradient: the negative of this reward-gradient estimate.
We use 8 samples per problem on a pool of 128 held-out problems, refreshed every 25 steps.

\section{Instantiation details and pseudocode}
\label{app:instantiation}

The coding instantiation uses the AZR training loop \citep{zhao2025absolutezero}.
The Lean instantiation adapts \sgs{} \citep{bailey2026sgs} to condition conjectures on solved seed theorems
(Section~\ref{sec:lean}).
Within each domain's setup, the methods share the training loop and solver objective.
They use different proposer rewards, or omit proposer optimization for \solveonly{}.
The coding baseline uses AZR's \difficulty{} reward.
The Lean baselines include the SGS guide-score-times-difficulty reward and \difficulty{}, which omits the guide.
Table~\ref{tab:instantiation} summarizes the two instantiations; Algorithm~\ref{alg:loop} gives one iteration.
In both domains, eligibility for the alignment reward requires $0 < s < 1$.

\begin{table}[h]
\centering
\caption{The two instantiations of the reward.}
\label{tab:instantiation}
\small
\setlength{\tabcolsep}{4pt}
\begin{tabular}{>{\raggedright\arraybackslash}p{0.22\linewidth}>{\raggedright\arraybackslash}p{0.35\linewidth}>{\raggedright\arraybackslash}p{0.35\linewidth}}
\toprule
 & coding self-play & theorem-proving self-play \\
\midrule
base loop & AZR \citep{zhao2025absolutezero} & adapted \sgs{} \citep{bailey2026sgs} \\
proposer / solver weights & shared, one model & separate copies of one base model \\
environment $v$ & Python execution & Lean~4 + mathlib4 \\
tasks & program-based deduction, abduction, induction & Lean statements \\
proposer conditioning $z$ & past self-generated tasks & a seed statement the solver has proved \\
verified solution $y^\star(\tilde{x})$ & oracle answer in the solver's format & up to $m = 4$ verified proofs, averaged \\
reference $\gref$ & external: mean gradient on 32 held-out MATH problems with gold solutions, fixed set, recomputed at current $\theta$ & local: gradient on the seed's own proof \\
proposer reward & $r_{\cos}$, replacing the difficulty reward for eligible candidates & $r_{\cos \times \mathrm{nov}}$, replacing the \sgs{} judge $\times$ difficulty reward; min-max normalized within the batch as in \sgs \\
eligible candidates & $0 < s < 1$ ($k = 8$) & $0 < s \le$ the batch's 70th solve-rate percentile ($k = 8$) \\
proposer update & REINFORCE++ with KL, jointly with the solver & reward-weighted REINFORCE \\
solver update & REINFORCE++ on binary correctness & REINFORCE on verified proofs of tasks with $s \le 1/2$ \\
where the reward is computed & inside the reward function, on the training workers & after verification, before the trainer starts, on one device \\
\bottomrule
\end{tabular}
\end{table}

\begin{algorithm}[h]
\caption{Self-play with a gradient-aligned proposer reward (one iteration).}
\label{alg:loop}
\begin{algorithmic}[1]
\REQUIRE solver $\pi_\theta$, proposer $\pi_\phi$, environment $v$, reference (external $\mathcal{D}_{\mathrm{ref}}$ or local seeds), samples per task $k$, proofs per candidate $m$ (local only)
\STATE candidates $\{\tilde{x}_i\} \leftarrow \pi_\phi(\cdot \mid z_i)$ \hfill $\triangleright$ propose
\FOR{each candidate and each target task $x$}
  \STATE sample $y_1, \ldots, y_k \sim \pi_\theta(\cdot \mid x)$; verify $v(x, y_j)$; $s(x) \leftarrow \frac{1}{k}\sum_j v(x, y_j)$ \hfill $\triangleright$ solve, verify
\ENDFOR
\IF{external reference}
  \STATE $\gref \leftarrow \nabla_\theta \frac{1}{|\mathcal{D}_{\mathrm{ref}}|}\sum_j \ell(x_j, y_j^\star; \theta)$ \hfill $\triangleright$ once per iteration
\ENDIF
\FOR{each eligible candidate $\tilde{x}_i$ ($0 < s(\tilde{x}_i) < 1$; Lean: $s(\tilde{x}_i) \le$ 70th percentile)}
  \STATE $y^\star(\tilde{x}_i) \leftarrow$ verified solution (oracle answer, or verified proofs)
  \STATE $g_i \leftarrow \nabla_\theta \ell(\tilde{x}_i, y^\star(\tilde{x}_i); \theta)$ \hfill $\triangleright$ no optimizer step
  \IF{local reference}
    \STATE $\gref \leftarrow \nabla_\theta \ell(x_{\mathrm{seed},i}, y^\star_{\mathrm{seed},i}; \theta)$
  \ENDIF
  \STATE $r_i \leftarrow \cos(g_i, \gref)$
  \IF{local reference}
    \STATE $r_i \leftarrow r_i \cdot \big(1 - J(\tilde{x}_i, x_{\mathrm{seed},i})\big)$ \hfill $\triangleright$ novelty factor
  \ENDIF
\ENDFOR
\STATE ineligible candidates keep the base loop's handling (penalty or exclusion)
\STATE update $\pi_\theta$ on verified attempts with reward $v$ \hfill $\triangleright$ base solver objective
\STATE update $\pi_\phi$ on candidates with reward $r$ \hfill $\triangleright$ base proposer objective
\end{algorithmic}
\end{algorithm}

\end{document}